\documentclass{ifacconf}

\usepackage{graphicx}      
\usepackage{natbib}        
\usepackage{graphics} 
\usepackage{amsmath} 
\usepackage{amssymb}  
\usepackage{xcolor}
\usepackage{subfig}
\usepackage{caption}
\usepackage{float}
\usepackage[vflt]{floatflt}

\DeclareMathOperator*{\argmin}{arg\,min}

\begin{document}
\begin{frontmatter}

\title{A novel Multiplicative Polynomial Kernel\\ for Volterra series identification} 


\author[First]{Alberto Dalla Libera, Ruggero Carli, Gianluigi Pillonetto} 

\address[First]{Authors are with the Department of Information Engineering, University of Padova, Via Gradenigo 6/B, 35131 Padova, Italy
	{\tt\small dallaliber@dei.unipd.it}, {\tt\small carlirug@dei.unipd.it}, {\tt\small giapi@dei.unipd.it}}

\begin{abstract}                
Volterra series are especially useful for nonlinear system identification, also thanks to their capability to approximate a broad range of input-output maps. However, their identification from a finite set of data is hard, due to the curse of dimensionality. Recent approaches have shown how regularization strategies can be useful for this task. In this paper, we propose a new regularization network for Volterra models identification. It relies on a new kernel given by the product of basic building blocks. Each block contains some unknown parameters that can be estimated from data using marginal likelihood optimization or cross-validation. In comparison with other algorithms proposed in the literature, numerical experiments show that our approach allows to better select the monomials that really influence the system output, much increasing the prediction capability of the model.
\end{abstract}


\end{frontmatter}

\section{Introduction}\label{sec:intro}
In many real world applications, linear models are not able to adequately describe dynamic systems.
This can be due to the presence of saturations, quantizers or static nonlinearities 
at the input and/or the output \cite{Ljung:99}[Section 5].
Even if some insight on the nonlinearities
can be available, the formulation of parametric models 
from finite data records is a difficult task \cite{Haber,Lind,Sjoberg:1995}. In particular, 
nonlinear system identification is often seen as an extended parametric regression
where the choice of regressors and basis functions plays a crucial role.
In this context, Volterra series are especially useful since they can represent a broad range of nonlinear systems \cite{Rugh1980,Boyd1985,Cheng2017}. When working in discrete-time, such models correspond to Taylor expansions of the input-output map. Indeed, a truncated Volterra series 
describes the system as the sum of all the possible monomials in the past inputs and outputs, up to 
a certain order. The problem is however the curse of dimensionality: 
the number of monomials grows quickly w.r.t. the polynomial degree and the system memory (given e.g. by the number of past input values that determine the output).
Thus, a careful selection of the relevant
components to be included in the model is crucial to control the complexity of the estimator, a problem known as regressors selection.

Suboptimal solutions are often searched through greedy approaches like
forward/backward subset selection, see for instance \cite{Chen1989,Billings1}. These regressor selection methods have however difficulties in handling high-dimensional regression spaces. An interesting option is joint estimation and variable selection. This can be performed using e.g. the $\ell_1$-norm regularizer which leads to the famous
LASSO \cite{Lasso1996}.\\
More recent approaches proposed to deal with the dimensionality of Volterra models can be found in \cite{Birpoutsoukis2017,Stoddard2017}. In particular, in \cite{Birpoutsoukis2017}, inspired by ideas developed for linear system identification in \cite{SS2010}, the authors proposed a regularization strategy suitable for Volterra series with smooth exponential decay.

An alternative route to the approaches mentioned above 
is the use of kernel-methods, which lead e.g. to the 
so called regularization networks \cite{Poggio90}.
Here, an unknown function is determined as the minimizer of 
an objective that is sum of two terms: a quadratic loss and
regularizer defined by a positive definite kernel. 
The choice of the kernel has a major effect on the quality of the estimate
since it encodes the expected properties of the function to reconstruct. 
Just looking at the function to reconstruct as the unknown system (input-output map), in recent years kernel-based approaches have been widely exploited also for nonlinear system identification 
and prediction, see for instance \cite{PartLin1,Hall2013}.\\
Another popular model is the polynomial kernel, which has a deep connection
with Volterra series. In fact, it implicitly encodes all the monomials 
up to the desired degree $r$, a kernel parameter tunable by the user.
Regularization networks for efficient Volterra identification that exploit this kernel can be found in \cite{Franz06aunifying}.

The approach described in this paper is based on a new polynomial kernel, named \emph{Multiplicative Polynomial kernel} (MPK). 
Similarly to the polynomial kernel, the MPK encodes all the monomials up to degree $r$, but it is defined by the product of $r$ linear kernels, each one equipped with a distinct set of hyperparameters. The MPK has some important features w.r.t. the polynomial kernel used in \cite{Franz06aunifying}. 
As already said, the polynomial kernel depends only on the polynomial degree $r$ and it encodes a number of monomials rapidly increasing with $r$ and the system memory. When plugged in a regularization network, it induces a penalty that cannot promote any sparsity in the solution. 
On the contrary, the MPK is equipped with a augmented set of hyperparameters, which allows promoting the monomials w.r.t. their maximum relative degree, improving regularization performance. 
Tests performed both in simulation and with data coming from a real system show that the MPK hyperparameters can be tuned via marginal likelihood optimization or cross-validation.

The paper is organized as follows. In Section \ref{sec:background} we provide a brief overview on Volterra series and the main identification approaches adopted. In Section \ref{sec:proposed_approach} we highlight some critic aspects of the standard polynomial kernel, and we introduce our kernel function, the Multiplicative Polynomial Kernel, highlighting its regularization capabilities. Finally, in Section \ref{sec:experimental_results} we report numerical results, in which we compare performance of the proposed kernel and the standard polynomial kernel.  

\section{BACKGROUND}\label{sec:background}

\subsection{Volterra series}
Let $u_k$ and $z_k$ be the one dimensional input and output signals at time $k$. When modeling the system response with a discrete time Volterra series of order $r$ the noisy output $y_k$ is assumed to be the sum of measurement noise and $r+1$ contributions acting on the lagged inputs $u_k,\, u_{k-1},\, u_{k-2}\,\dots$. Assume that the system has finite memory $m$, and define the input vector $\boldsymbol{u}_k = [u_k \, \dots \, u_{k-m}]$. Then we have
\begin{align}
y_k &= z_k + e_k = h_0 + \sum_{i=1}^{r}H_i(\boldsymbol{u}_k) + e_k \label{eq:volterra_kernel} \text{ ,}
\end{align}
where $e_k \sim N\left(0,\sigma_n^2\right)$ is the measurement noise, $h_0$ is a constant accounting for the zero-order Volterra contribution, while the $H_i$ are the \emph{i}-th Volterra contributions. In particular, each $H_i$ is defined as the convolution between $\boldsymbol{u}_k$ and a Volterra map $h_i$, namely,
\begin{align}\label{eq:VolterraKernel}
H_i(\boldsymbol{u}_k) = &\sum_{\tau_1=0}^{m} \dots \sum_{\tau_i=0}^{m} h_i\left(\tau_1,\dots,\tau_i\right)\prod_{\tau=\tau_1}^{\tau_i}u_{k-\tau}.
\end{align}
In this paper, we consider symmetric Volterra kernels, i.e., given a set of lags $\tau_1,\dots,\tau_i$, the value of $h_i$ is independent on the lags order. For instance, with $i=2$ we have $h_2\left(\tau_1,\tau_2\right)=h_2\left(\tau_2,\tau_1\right)$.\\
Alternatively, each $H_i$ term in \eqref{eq:volterra_kernel} can be rewritten more compactly as an inner product. Let $\boldsymbol{\phi}_i(\boldsymbol{u}_k)$ be a vector collecting all the distinct monomials with degree $i$ in the components of $\boldsymbol{u}_k$. We have 
\begin{equation*}
H_i(\boldsymbol{u}_k) = \boldsymbol{\phi}^T_i\left(\boldsymbol{u}_{k}\right) \boldsymbol{w}_i \text{,}
\end{equation*}
where $\boldsymbol{w}_i$ is the vector collecting the distinct $h_i$ coefficients ordered in accordance with $\boldsymbol{\phi}_i$, and opportunely scaled to account for repetitions due to symmetry. More precisely, the $\boldsymbol{w}_i$ entry associated to the monomial $\prod_{j=0}^{m} u_{k-j}^{d_j}$ is the product between the correspondent $h_i$ coefficient scaled by the multinomial coefficient $\binom{i}{d_0,\dots,d_m}$. 
Based on the above definitions, \eqref{eq:volterra_kernel} can be rewritten as
\begin{equation}
y_k = \boldsymbol{\phi}\left(\boldsymbol{u}_k\right)^T\boldsymbol{w} + e_k \label{eq:volterra_linear} \text{,}
\end{equation}
where 
\begin{align}
&\boldsymbol{\phi}\left(\boldsymbol{u}_k\right) = \left[h_0\,\boldsymbol{\phi}^T_1(\boldsymbol{u}_k)\,\dots\, \boldsymbol{\phi}^T_r(\boldsymbol{u}_k)\right]^T \text{,} \label{eq:phi_volterra}\\
&\boldsymbol{w}\left(\boldsymbol{u}_k\right) = \left[w_0\,\boldsymbol{w}^T_1\,\dots\, \boldsymbol{w}^T_r\right]^T \text{.}
\end{align}

The authors in \cite{Birpoutsoukis2017} have proposed to learn the input-output relations by directly estimating the elements of $\boldsymbol{w}$. This estimation is performed by solving a least square problem defined by \eqref{eq:volterra_linear}, given a data set of input-output measurements $D = \left\{\left(\boldsymbol{u}_k,y_k\right), \, k=1,\dots T \right\}$.\\
It is worth stressing that the applicability of algorithms based on a least-square approach is strongly limited by the high computational and memory requirements related to the dimension of $\boldsymbol{w}$. Indeed, the number of Volterra coefficients grows rapidly with the system memory $m$ and the Volterra order $r$. More precisely, assuming that the Volterra maps are symmetric, we have that $\boldsymbol{w}_i$ is composed of $N_i = \binom{m + i}{i}$ elements, leading to a total number of $N =  1 + \sum_{i=1}^{r}N_i$ parameters to be estimated.

\subsection{Polynomial kernel and Volterra series}

An alternative solution to accomplish the Volterra series identification has been proposed in \cite{Franz06aunifying}. Instead of formulating the identification problem directly w.r.t. $\boldsymbol{w}$, the authors rely on kernel based techniques. The input-output map $f$ is assumed to belong to a \emph{reproducing kernel Hilbert space} (RKHS) \cite{learning_with_kernels}, defined by a kernel function $k(\boldsymbol{u}_k, \boldsymbol{u}_j)$. Given an input-output dataset $D$ like that previously introduced, $\hat{f}$, the estimate of $f$, is obtained solving the following problem,
\begin{equation}
\argmin_{f \in H} \sum_{t=1}^{T}\left(y_t-z(\boldsymbol{u}_t)\right)^2 + \gamma^2||f||_H^2 \text{,} \label{eq:loss}
\end{equation}
where the first term of the loss function accounts for the adherence to experimental data, while the second is the regularization term, given by the squared RKHS norm of $f$; the balance between these two contributions can be controlled tuning the hyperparameter $\gamma$. According to the \emph{representer theorem}, $\hat{f}$ is expressed in closed form as
\begin{equation}\label{eq:f_hat}
\hat{f}(\boldsymbol{u}_k) = \hat{z}_k = \sum_{t=1}^{T} \alpha_t k(\boldsymbol{u}_k, \boldsymbol{u}_t) \text{,}
\end{equation}
where $\boldsymbol{\alpha} = \left[\alpha_1,\dots, \alpha_T\right]^T$ is equal to $(K+ \gamma^2I_T)^{-1}\boldsymbol{y}$,
$\boldsymbol{y} = \left[y_1,\dots,y_T\right]^T$ denotes the vector containing all the output measurements, and $K$ is the Kernel matrix, i.e. its $(k,j)$ entry is $K_{k,j}=k(\boldsymbol{u}_k, \boldsymbol{u}_j)$.\\
For our future use, it is also useful recalling the following fundamental facts regarding RKHS theory. Under mild assumptions, a kernel function admits an expansion (possibly infinite) in terms of
basis functions $\phi_q$, namely,
\begin{align}
&k(\boldsymbol{u}_k, \boldsymbol{u}_j) = \sum_{q} \lambda_q \phi_q(\boldsymbol{u}_k)\phi_q(\boldsymbol{u}_j) \text{,} \label{eq:poly_expansion}
\end{align}
where $\lambda_q$ are positive scalars. It can then be proved that any function in the RKHS induced by the above kernel has the representation
\begin{equation}
f(\boldsymbol{u}_k) = \sum_{q}c_q \phi_q(\boldsymbol{u}_k) \text{,}
\end{equation}
for suitable coefficients $c_q$. In addition, if all the basis functions $\phi_q$ are linearly independent, one also has
\begin{equation}
||f||_H^2 = \sum_{q}\frac{c^2_q }{\lambda_q} \text{.} \label{eq:penalty}
\end{equation}
This last relation shows how the $\lambda_q$ coefficients are related to each $\phi_q$ in determining the regularization term present in (\ref{eq:loss}). In particular, small values of $\lambda_q$ entail high penalization of $\phi_q$.

As far as the kernel function is concerned, in \cite{Franz06aunifying} the authors considered the polynomial kernel. In particular, we discuss the inhomogeneous polynomial kernel, defined as
\begin{equation}
k^{(r)}(\boldsymbol{u}_k, \boldsymbol{u}_j) = \left( 1 + \boldsymbol{u}_k^T  \boldsymbol{u}_j\right)^r \text{,} \label{eq:poly_kernel}
\end{equation}
where $k$ is a tunable hyperparameter corresponding to the degree of the polynomial kernel. As showed in \cite{learning_with_kernels}, the polynomial kernel in \eqref{eq:poly_kernel} admits an expansion in the monomials in $\boldsymbol{u}_k$, with degree up to $k$. Namely, referring to \eqref{eq:poly_expansion} and \eqref{eq:penalty}, we have that the $\phi_q$ corresponds to the elements of the $\boldsymbol{\phi}$ vector defined in \eqref{eq:phi_volterra}; accordingly, $1/\lambda_q$ then defines the penalty assigned to the relative monomial.

We conclude this subsection with a computational note. The computation of $\hat{f}$ involves the inversion of a $T \times T$ matrix, see \eqref{eq:f_hat}. The number of operations so scales with the cube of the data set size, and there is no direct dependence on $N$, the dimensions of $\boldsymbol{\phi}$, allowing the use of high-order Volterra models.

\section{PROPOSED KERNEL}\label{sec:proposed_approach}
The Volterra series learning strategy we propose is based on a novel polynomial kernel, called \emph{Multiplicative Polynomial Kernel} (MPK). Compared to the standard polynomial kernel reported in \eqref{eq:poly_kernel}, our kernel is equipped with a set of parameters which allows to assign suitable priors to the different basis functions of the RKHS, thus leading to better performance in terms of estimation and generalization. Before describing the proposed kernel function we highlight some critical issues of standard inhomogeneous polynomial kernel.

\subsection{Penalties induced by \eqref{eq:poly_kernel}}
As stated in \cite{Rasmussen} (Chapter 4.2.2), polynomial kernels are not widely used in regression problems, since they are prone to overfitting, in particular, in presence of high dimensional inputs and when the degree is greater than two. Indeed, in the kernel formulation given in \eqref{eq:poly_kernel} there are not parameters that allow to weigh differently the monomials composing the RKHS. 

To clarify this concept we consider a simple example, a third order Volterra series with $m = 1$ defined as follow
\begin{equation}
f(\boldsymbol{u}_k) = u^3_{k} + u^2_{k}u_{k-1} +0.5 \label{eq:test_case} \text{.}
\end{equation}
We compute the $\lambda_q$ obtained with the standard polynomial kernel expanding \eqref{eq:poly_kernel}, and comparing the result with \eqref{eq:poly_expansion}. For the sake of clarity, we will denote with $\lambda_{d_0,\dots,d_{m}}$ the penalty coefficient associated to the monomial $\prod_{\tau=0}^{m}u^{d_{\tau}}_{k-\tau}$. The kernel expansion is  
\begin{align*}
k^{(3)}(\boldsymbol{u}_i,\boldsymbol{u}_j) &= u_{i}^3u_{j}^3 + u_{i-1}^3u_{j-1}^3 \\
&+ 3 u_{i}^2u_{i-1}u_{j}^2u_{j-1} + 3 u_{i}u_{i-1}^2u_{j}u^2_{j-1} \\
&+ 3u_{i}^2u_{j}^2 + 3u_{i-1}^2u_{j-1}^2 + 6u_{i}u_{i-1}u_{j}u_{j-1} \\
&+ 3u_{i}u_{j} + 3u_{i-1}u_{j-1} + 1 \text{.}
\end{align*}
Then, by inspection, we obtain
\begin{align*}
&\lambda_{3,0} = \lambda_{0,3} = 1 \text{, } \lambda_{2,1} = \lambda_{1,2} = 3 \text{,}\\
&\lambda_{2,0} = \lambda_{0,2} = 3 \text{, } \lambda_{1,1} = 6 \text{,}\\
&\lambda_{1,0} = \lambda_{0,1} = 3 \text{, } \lambda_{0,0} = 1  \text{.}
\end{align*}    
These values show that \eqref{eq:poly_kernel} assigns penalties based on the monomial degree, and promoting mixed terms. This trend might not be representative of the Volterra kernel, leading to the need of more training data to obtain accurate estimates. For instance, consider the function reported in \eqref{eq:test_case}. It is evident that the $\lambda$ values obtained with \eqref{eq:poly_kernel} do not describe properly the contributions of the different monomials, since, for example, the higher values of $\lambda$ are assigned to monomials that are not present.

\subsection{Multiplicative Polynomial Kernel}
The kernel function we propose to model the $r$ order Volterra series is given by the product of $r$ linear kernels, and it is formally defined as
\begin{equation}
k^{(r)}(\boldsymbol{u}_k, \boldsymbol{u}_j) = \prod_{i=1}^{r} \left(\sigma_{0_i} +( \boldsymbol{u}_k)^T \Sigma_i \boldsymbol{u}_j\right) \text{,} \label{eq:proposed_kernel}
\end{equation}
where the matrices $\Sigma_i \in \mathbb{R}^{(m+1) \times (m+1)}$ are diagonal. In particular, for each $i$ we have $\Sigma_i = diag \left( \left[\sigma^{(i)}_{0},\dots,\sigma^{(i)}_{m}\right] \right)$, with the diagonal elements greater of equal than zero.\\
Exploiting the kernel properties it can be easily shown that the function defined in \eqref{eq:proposed_kernel} is a well-defined kernel function, since it is the product of several valid kernel functions, see \cite{Rasmussen}.

\subsection{Penalties induced by the MPK}
In this subsection we analyze the advantages of the proposed kernel function, focusing in the role played by the kernel parameters. To this aim, we consider the example analyzed in the previous subsection, that is, the identification of the input-output behavior of a Volterra series with $r=3$ and $m=1$. Starting from the kernel definition given in \eqref{eq:proposed_kernel}, through standard algebraic computations, we can derive the penalties coefficients as functions of the MPK parameters. In particular, the penalties assigned to monomials of degree three are
\begin{align*}
&\lambda_{3,0} = \prod_{j=1}^{3}\sigma^{(j)}_{0}\text{,}\,\,\,\, \lambda_{0,3} = \prod_{j=1}^{3}\sigma^{(j)}_{1} \text{,} \\
&\lambda_{2,1} = \sum_{j=1}^{3} \sigma^{(j)}_{1} \prod_{l \neq j}\sigma^{(l)}_{0}\text{,}\,\,\,\, \lambda_{1,2} = \sum_{j=1}^{3} \sigma^{(j)}_{0} \prod_{l \neq j}\sigma^{(l)}_{1} \text{,}
\end{align*}
the ones assigned to monomials of degree two are
\begin{align*}
&\lambda_{2,0} = \sum_{j=1}^{3}\sigma_{0_j}\prod_{l \neq j}\sigma^{(l)}_{0}\text{,}\,\,\,\, \lambda_{0,2} = \prod_{j=1}^{3}\sigma_{0_j}\prod_{l \neq m}\sigma^{(l)}_{1} \text{,} \\
&\lambda_{1,1} =  \sum_{j=1}^{3} \sigma_{0_j} \sum_{l_1\neq l_2 \neq j}\sigma^{(l_1)}_{0} \sigma^{(l_2)}_{1}\text{,}
\end{align*}
and, finally, the ones assigned to monomials of degree one and zero are
\begin{align*}
&\lambda_{1,0} = \sum_{j=1}^{3}\sigma^{(j)}_{0}\prod_{l \neq j}\sigma_{0_l}\text{,}\,\,\,\, &\lambda_{0,1} = \sum_{j=1}^{3}\sigma^{(j)}_{1}\prod_{l \neq j}\sigma_{0_l}\text{,} \\
&\lambda_{0,0} = \prod_{j=1}^{3}\sigma_{0_j}\text{,}
\end{align*}
Some interesting insights can be obtained from the previous penalties expressions. 
Notice that the MPK parameters allow penalizing the monomials w.r.t. their relative degree. For instance, consider the penalties assigned to monomials that contains $u_k$, namely, $\lambda_{3,0}$, $\lambda_{2,1}$, $\lambda_{1,2}$, $\lambda_{2,0}$, $\lambda_{1,1}$, and $\lambda_{1,0}$, as function of $\sigma^{(j)}_{0}$, with $j = 1,2,3$. Analyzing these expression we can appreciate that to promote monomials in which $u_k$ appears with degree \emph{i} at least $i$ of the $\lambda^{(j)}_0$ parameters need to be significantly greater than zero. On the contrary, to penalize monomials in which $u_k$ appears with relative degree greater than $i$ we need to set to zero $r-i$  of the $\sigma^{(j)}_{0}$ elements. Thus, referring to the test function in \eqref{eq:test_case}, where the maximum relative degree of $u_k$ and $u_{k-1}$ are, respectively, $3$ and $1$, we can exclude part of the monomials that do not influence the input-output defining
\begin{align}
&\sigma_{0_1} = 1 \text{ , } \left[ \sigma^{(1)}_{0} \, \sigma^{(1)}_{1} \right] = \left[1 \,1\right]\text{,} \nonumber\\
&\sigma_{0_2} = 1 \text{ , } \left[ \sigma^{(2)}_{0} \, \sigma^{(2)}_{1} \right] = \left[1 \,0\right]\text{,}\nonumber\\
&\sigma_{0_3} = 1 \text{ , } \left[ \sigma^{(3)}_{0} \, \sigma^{(3)}_{1} \right] = \left[1 \,0\right]\text{,} \label{eq:test_case_parameters}
\end{align}
which determine the following penalties,
\begin{align}
&\lambda_{3,0} = 1\text{, } \lambda_{2,1} = 1 \text{, }  \lambda_{0,3} = \lambda_{1,2} = 0\text{,}\nonumber\\
&\lambda_{2,0} = 2\text{, } \lambda_{0,2} = 0 \text{, } \lambda_{1,1} = 2 \text{,}\nonumber\\
&\lambda_{1,0} = 2\text{, } \lambda_{0,1} = 1 \text{, } \lambda_{0,0} = 1  \text{.} \label{eq:test_case_lambda}
\end{align}    
The hyperparameters tuning can be accomplished relying on empirical methods, like cross validation, or optimizing a given loss function, like \emph{Marginal Likelihood} (ML).

\subsection{Parametrization of the $\Sigma_i$ matrices}
Notice that the MPK is given by the product of $r$ equal blocks. Consequently, permuting the $\Sigma_i$ matrices we obtain different configurations of the MPK hyperparameters associated to the same set of penalties. For instance, referring to the test case analyzed in the previous subsection, the configuration reported in \eqref{eq:test_case_parameters} and the following configuration are associated to the same set of penalties, i.e. \eqref{eq:test_case_lambda}, 
\begin{align*}
&\sigma_{0_1} = 1 \text{ , } \left[ \sigma^{(1)}_{0} \, \sigma^{(1)}_{1} \right] = \left[1 \,0\right]\text{,}\\
&\sigma_{0_2} = 1 \text{ , } \left[ \sigma^{(2)}_{0} \, \sigma^{(2)}_{1} \right] = \left[1 \,1\right]\text{,}\\
&\sigma_{0_3} = 1 \text{ , } \left[ \sigma^{(3)}_{0} \, \sigma^{(3)}_{1} \right] = \left[1 \,0\right]\text{.} 
\end{align*}

When optimizing the hyperparameters by ML, this fact could lead to undesired behaviors, due to the presence of several local maxima. To avoid such behaviors, we propose a iterative parametrization of the $\Sigma_i$ diagonal elements. More specifically, the $\Sigma_i$, are defined by a backward iteration as follows, 
\begin{align}
\Sigma_r &= diag \left(\left[a^{(r)}_{0},\dots,a^{(r)}_m\right]\right) \text{,}  \label{eq:kernel_diag}\\
\Sigma_i &= \Sigma_{i+1} + diag \left(\left[a^{(i)}_{0},\dots,a^{(i)}_m\right]\right) \text{,} \notag 
\end{align}
where the $a^{(i)}_j$ elements are greater of equal than zero. We conclude by emphasizing the relation between the proposed re-parameterization and the relative degree with which each term appears. Notice that increasing $a^{(i)}_j$ we promote simultaneously all the monomials in which $u_{k-j}$ appears with relative degree up to $i$. Moreover, to penalize monomials in which $u_{k-j}$ appears with relative degree greater than $i$ we need to set to zero the $a^{(l)}_{j}$ with $l > i$.


\begin{figure*}[]
	\begin{minipage}[h!]{.5\textwidth}
		\centering
		\includegraphics[width=1\linewidth]{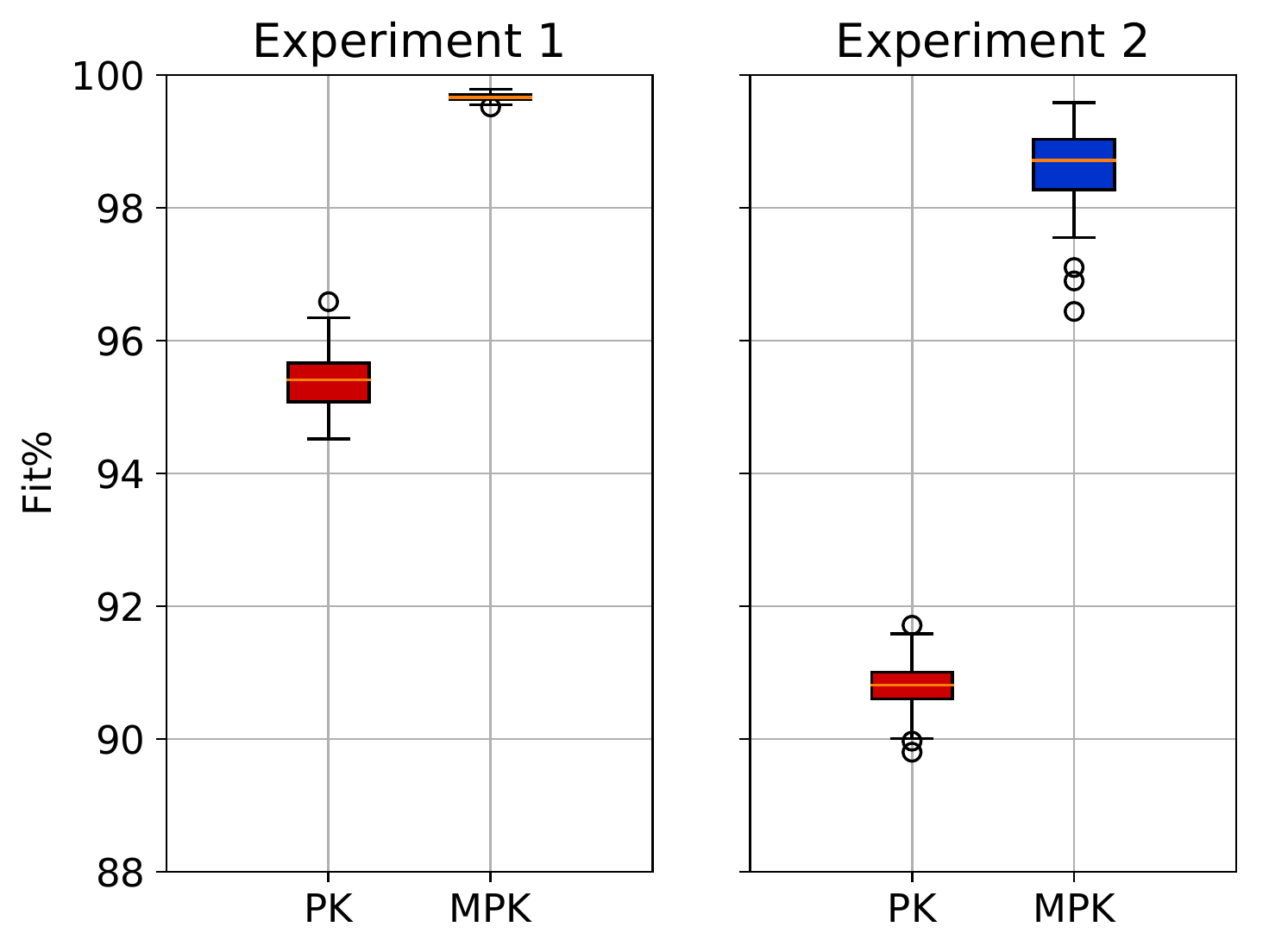}
	\end{minipage}
	\hfill
	\begin{minipage}[h!]{.5\textwidth}
		\centering
		\includegraphics[width=1\linewidth]{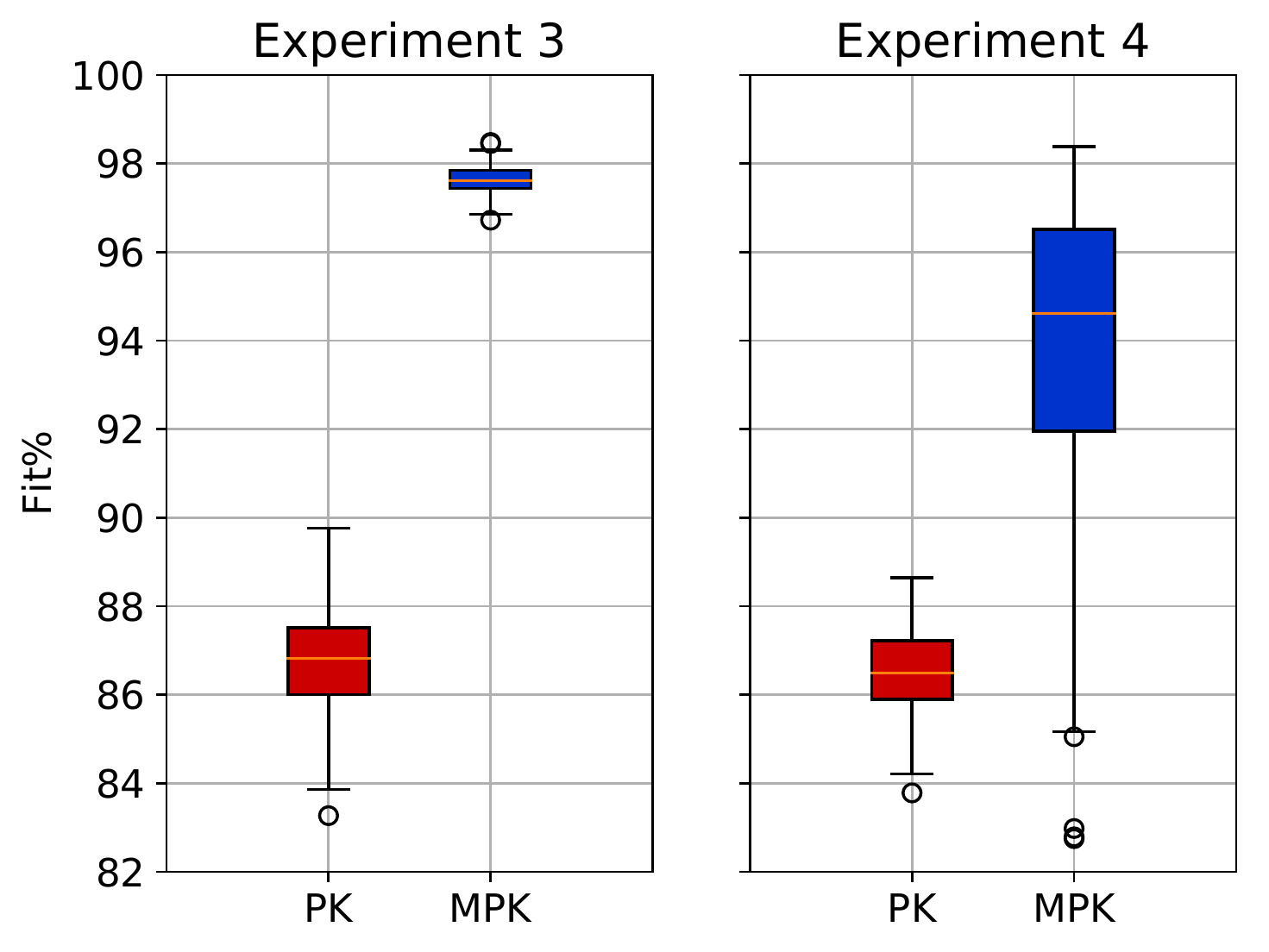}
	\end{minipage}  
	\caption{Boxplots of the 100 test set Fit\% obtained in the different scenarios considered.}
	\label{fig:boxplots}
\end{figure*}


\section{EXPERIMENTAL RESULTS} \label{sec:experimental_results}
\subsection{Simulated environment}
In this set of experiments we test the performance of the MPK in a simulated environment, the benchmark system introduced in \cite{SPL_system}, i.e. a third order Volterra series described by the following equation
\begin{align}
z_k &= u_k + 0.6 u_{k-1} + 0.35(u_{k-2}+u_{k-4}) -0.25u^2_{k-3} \nonumber \\
&+ 0.2(u_{k-5}+u_{k-6}) + 0.9 u_{k-3} + 0.25u_{k}u_{k-1} +0.75u^3_{k-2}\nonumber\\
&-u_{k-1}u_{k-2} +0.5(u^2_{k}+u_{k}u_{k-2} + u_{k-1}u_{k-3}) \text{.} \label{eq:SPL}
\end{align}
The MPK-based estimator is compared with the one based on the polynomial kernel reported in \eqref{eq:poly_kernel}, hereafter denoted with PK; both for Pk and MPK we considered $r=3$. Input signals are $1000$ samples obtained from a realization of Gaussian noise. Concerning $m^{tr}_{u}$, $m^{ts}_{u}$, $\sigma^{tr}_{u}$ and $\sigma^{ts}_{u}$, respectively, the input mean and standard deviation of the training and test samples, we consider four different scenarios: 
\begin{itemize}
	\item \textit{Experiment 1}: $m^{tr}_{u}=m^{ts}_{u}=0$, $\sigma^{tr}_{u}=\sigma^{ts}_{u}=4$;
	\item \textit{Experiment 2}: $m^{tr}_{u}=m^{ts}_{u}=0$, $\sigma^{tr}_{u}=\sigma^{ts}_{u}=2$;
	\item \textit{Experiment 3}: $m^{tr}_{u}=-12$, $m^{ts}_{u}=12$, $\sigma^{tr}_{u}=\sigma^{ts}_{u}=4$;
	\item \textit{Experiment 4}: $m^{tr}_{u}=-12$, $m^{ts}_{u}=12$, $\sigma^{tr}_{u}=\sigma^{ts}_{u}=2$.
\end{itemize}
In all the experiments the noise standard deviation is $\sigma_n=4$. The PK and MPK hyperparameters have been trained optimizing the ML of the training samples. As concerns the optimization, we used standard gradient descent algorithm, with adaptive learning rate. The two estimators are implemented in PyTorch \cite{pytoch}, to exploit automatic differentiation functionalities.\\
The four experiments can be grouped in two sets. Generalization properties are stressed more in \textit{Experiment 3} and \textit{Experiment 4} since $m^{tr}_{u} \neq m^{ts}_{u}$, and hence the training and test input signals are significantly different with each other; in particular, the mean values are such that with high probability the test inputs are outside the $3\sigma$ of training inputs distribution. In \textit{Experiment 1} (resp. \textit{Experiment 3}) and \textit{Experiment 2} (resp. \textit{Experiment 4}) we considered different values of the input standard deviation in order to analyze the estimators behaviors with different system excitations.\\
For each experiment we performed a Monte Carlo of 100 simulations. In each simulation the same training and test data sets have been used to implement and test the MPK and PK based estimators. Results are reported in Figure \ref{fig:boxplots}. Performance is measured by the percentage fit (Fit\%), defined as
\begin{equation*}
100\%\left(1-\frac{||\boldsymbol{z}-\hat{\boldsymbol{z}}||_1}{||\boldsymbol{z}-\bar{z}||_1}\right) \text{,}
\end{equation*}
where $\boldsymbol{z}$ and $\hat{\boldsymbol{z}}$ are the vectors collecting the real and estimated system output, while $\bar{z}$ is the mean of $\boldsymbol{z}$.

In Figure \ref{fig:boxplots}, we visualized the results obtained through some boxplots. The estimator based on the MPK outperforms the standard polynomial kernel, since in all the tests the Fit\% obtained with MPK are higher than the ones obtained with PK. Besides improving the estimation accuracy, the MPK parametrization improves also the generalization performance. Indeed, comparing results obtained in \textit{Experiment 1} and \textit{Experiment 3}, we can appreciate how the penalties learned by the MPK estimator by optimizing ML provides robustness to variations of the inputs distribution; more specifically, the MPK performance decreases less than the PK performance when the test input locations that are far from the training inputs.\\
As far as variations of the system excitation, comparing results obtained in \textit{Experiment 3} and \textit{Experiment 4}, we can observe how not sufficiently exciting training samples can lead to a bad identification of the MPK parameters. Notice how from \textit{Experiment 3} to \textit{Experiment 4}, the variance of MPK based estimator grows up more that the one of the PK based, highlighting the importance of using sufficiently exciting input trajectories.

\subsection{Identification of the Silverbox system}
\begin{figure}[]
	\centering
    \includegraphics[width=1\linewidth]{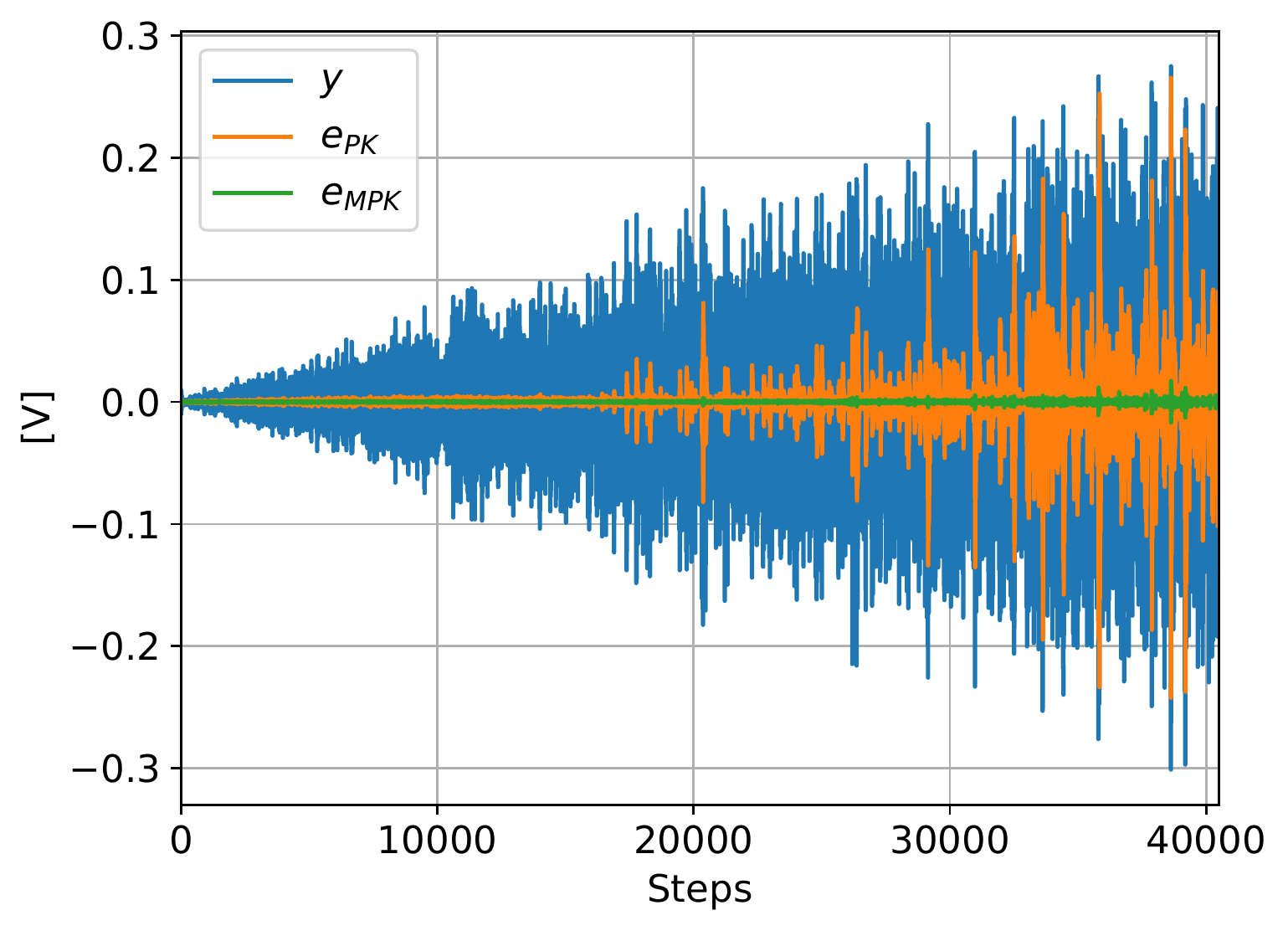}
	\caption{evolution of the test output together with the simulation errors $e_{PK}$ and $e_{MPK}$.}
	\label{fig:silverbox_sim_error}
\end{figure}

In this subsection we test the estimators based on MPK and PK with data collected on a real system, the Silverbox \cite{silverbox}. The Silverbox is an electrical system that simulates a mass-spring-damper system. The spring exhibits non-linear behaviors, and the system is described by the following differential equation
\begin{equation*}
m \ddot{z}(t) + d \dot{z}(t) + k_1 z(t) + k_3 z^3(t) = u(t) \text{,}
\end{equation*}
where $z$ and $u$ are, respectively, the mass displacement and the input force applied to the mass, while $d$, $k_1$, and $k_3$ are the parameters of the damper and the non-linear spring.\\
We used the MPK and PK based estimators to learn the evolution of the mass displacement, modeling the unknown target function with a third order Volterra series. We considered as input of the model the past $m=5$ $u$ and $y$, namely, at time $k$, the model input and output are, respectively, $\left[ u_k\, \dots u_{k-m}\, z_{k-1}\, \dots\, z_{k-m} \right]$ and $z_k$. The original training and test dataset are publicly available\footnote{http://www.nonlinearbenchmark.org/}. The training dataset accounts approximately for $80000$, obtained inputting to the system an odd random phase multisine signal, while the test set accounts for approximately $40000$ samples, collected exciting the system with filtered Gaussian noise.\\
To further stress generalization properties, we derived and trained the estimators using just the first $200$ samples of the training dataset. Besides testing the one-step-ahead prediction performance, we measured also the simulation performance: given an initial state of the system we simulate its evolution evaluating iteratively the function learned for the one-step-ahead problem, inputting to the estimator the past predicted output instead of the measured output. In this context, we noticed that optimizing the hyperparameters through cross-validation increases the simulation performance. To deal with the considerable number of parameters, we adopted a gradient-based strategy. Specifically, we randomly select $5$ partitions of the training data. Each partition accounts for two sets, composed of $100$ samples. As loss loss function we considered the sum of the mean squared errors (MSEs) in validation. Namely, for each partition, we derive the estimator based on the first set, and we compute the MSE of the second set. The loss function is the sum of the MSEs. The kernel hyperparameters are updated minimizing the loss though a gradient-based procedure, iterated until convergence of the loss.\\  
Performance is reported in Table \ref{tab:silverbox}. As before, we compare the one-step-ahead performance using the Fit\%. Simulation performance are measured both with Fit\% and root mean squared error (RMSE). The MPK outperforms the standard polynomial kernel, both in one-step-ahead prediction and simulation. The gap is particularly evident in simulation, where MPK significantly outperforms PK; see Figure \ref{fig:silverbox_sim_error} to compare the evolution of the simulation errors. Despite we used just the first $200$ training samples to derive the model, and hyperparameters optimization was focused on optimizing the one-step-ahead performance, the MPK performance is close ot the best results obtained in this benchmark.

\begin{table}[]
	\captionsetup{width=0.45\textwidth}
	\centering
	\caption{One-step-ahead (Pred.) and simulation (Sim.) performance of the PK and MPK based estimators obtained in the Silverbox test dataset.}
	\begin{tabular}{ |c|c|c|c| } 
		\hline
		 & Pred. (Fit\%)& Sim. (Fit\%) & Sim. (RMSE [mV]) \\
		\hline
		PK & $97.79$ & $81.13$ & $17.2213$\\
		\hline
		MPK & $99.70$ & $98.67$ & $0.8862$\\
		\hline
	\end{tabular}
    \label{tab:silverbox}
\end{table}

\section{CONCLUSIONS}
In this paper we have introduced a novel polynomial kernel, name \emph{Multiplicative Polynomial Kernel}. Compared to the standard polynomial kernel, the MPK is equipped with a set of parameters that allows to better select the monomials that really influence the system output. As proven by numerical results, this fact entails a performance improvement, both in terms of accuracy and generalization properties. 

\bibliography{references,biblio,biblioIdPWA,biblioSurvey,biblioWiener}

\begin{thebibliography}{22}
\providecommand{\natexlab}[1]{#1}
\providecommand{\url}[1]{\texttt{#1}}
\providecommand{\urlprefix}{URL }
\expandafter\ifx\csname urlstyle\endcsname\relax
  \providecommand{\doi}[1]{doi:\discretionary{}{}{}#1}\else
  \providecommand{\doi}{doi:\discretionary{}{}{}\begingroup
  \urlstyle{rm}\Url}\fi

\bibitem[{Billings et~al.(1989)Billings, Chen, and Korenberg}]{Billings1}
Billings, S., Chen, A., and Korenberg, M. (1989).
\newblock Identification of {MIMO} non-linear systems using a
  forward-regression orthogonal algorithm.
\newblock \emph{Intern. J. of Control}, 49, 2157 -- 2189.

\bibitem[{Birpoutsoukis et~al.(2017)Birpoutsoukis, Marconato, Lataire, and
  Schoukens}]{Birpoutsoukis2017}
Birpoutsoukis, G., Marconato, A., Lataire, J., and Schoukens, J. (2017).
\newblock Regularized nonparametric volterra kernel estimation.
\newblock \emph{Automatica}, 82, 324 -- 327.

\bibitem[{{Boyd} and {Chua}(1985)}]{Boyd1985}
{Boyd}, S. and {Chua}, L. (1985).
\newblock Fading memory and the problem of approximating nonlinear operators
  with volterra series.
\newblock \emph{IEEE Transactions on Circuits and Systems}, 32(11), 1150--1161.

\bibitem[{Chen et~al.(1989)Chen, Billings, and Luo}]{Chen1989}
Chen, S., Billings, S.A., and Luo, W. (1989).
\newblock Orthogonal least squares methods and their application to non-linear
  system identification.
\newblock \emph{International Journal of Control}, 50, 1873--1896.

\bibitem[{Cheng et~al.(2017)Cheng, Peng, Zhang, and Meng}]{Cheng2017}
Cheng, C., Peng, Z., Zhang, W., and Meng, G. (2017).
\newblock Volterra-series-based nonlinear system modeling and its engineering
  applications: A state-of-the-art review.
\newblock \emph{Mechanical Systems and Signal Processing}, 87, 340 -- 364.

\bibitem[{Espinoza et~al.(2005)Espinoza, Suykens, and {De Moor}}]{PartLin1}
Espinoza, M., Suykens, J.A.K., and {De Moor}, B. (2005).
\newblock Kernel based partially linear models and nonlinear identification.
\newblock \emph{IEEE Trans. on Automatic Control}, 50(10), 1602--1606.

\bibitem[{Franz and Sch\"olkopf(2006)}]{Franz06aunifying}
Franz, M. and Sch\"olkopf, B. (2006).
\newblock A unifying view of {W}iener and volterra theory and polynomial kernel
  regression.
\newblock \emph{Neural Computation}, 18, 3097--3118.

\bibitem[{Haber and Unbehauen(1990)}]{Haber}
Haber, R. and Unbehauen, H. (1990).
\newblock Structure identification of nonlinear systems-a survey.
\newblock \emph{Automatica}, 26, 651--677.

\bibitem[{Hall et~al.(2012)Hall, Rasmussen, and Maciejowski}]{Hall2013}
Hall, J., Rasmussen, C., and Maciejowski, J. (2012).
\newblock Modelling and control of nonlinear systems using {G}aussian processes
  with partial model information.
\newblock In \emph{Proceedings of the 51st Annual Conference on Decision and
  Control (CDC)}.

\bibitem[{Lind and Ljung(2008)}]{Lind}
Lind, I. and Ljung, L. (2008).
\newblock Regressor and structure selection in {{NARX}} models using a
  structured {ANOVA} approach.
\newblock \emph{Automatica}, 44, 383--395.

\bibitem[{Ljung(1999)}]{Ljung:99}
Ljung, L. (1999).
\newblock \emph{System Identification - Theory for the User}.
\newblock Prentice-Hall, Upper Saddle River, N.J., 2nd edition.

\bibitem[{Paszke et~al.(2017)Paszke, Gross, Chintala, Chanan, Yang, DeVito,
  Lin, Desmaison, Antiga, and Lerer}]{pytoch}
Paszke, A., Gross, S., Chintala, S., Chanan, G., Yang, E., DeVito, Z., Lin, Z.,
  Desmaison, A., Antiga, L., and Lerer, A. (2017).
\newblock Automatic differentiation in pytorch.

\bibitem[{Pillonetto and {De Nicolao}(2010)}]{SS2010}
Pillonetto, G. and {De Nicolao}, G. (2010).
\newblock A new kernel-based approach for linear system identification.
\newblock \emph{Automatica}, 46(1), 81--93.

\bibitem[{Poggio and Girosi(1990)}]{Poggio90}
Poggio, T. and Girosi, F. (1990).
\newblock {N}etworks for approximation and learning.
\newblock In \emph{Proceedings of the {IEEE}}, volume~78, 1481--1497.

\bibitem[{Rasmussen and Williams(2006)}]{Rasmussen}
Rasmussen, C. and Williams, C. (2006).
\newblock \emph{{G}aussian Processes for Machine Learning}.
\newblock The MIT Press.

\bibitem[{Rugh(1980)}]{Rugh1980}
Rugh, W. (1980).
\newblock \emph{Nonlinear System Theory: The Volterra-Wiener Approach}.
\newblock Johns Hopkins University Press.

\bibitem[{Scholkopf and Smola(2001)}]{learning_with_kernels}
Scholkopf, B. and Smola, A.J. (2001).
\newblock \emph{Learning with Kernels: Support Vector Machines, Regularization,
  Optimization, and Beyond}.
\newblock MIT Press, Cambridge, MA, USA.

\bibitem[{Sj\"{o}berg et~al.(1995)Sj\"{o}berg, Zhang, Ljung, A.~Benveniste,
  Glorennec, Hjalmarsson, and Juditsky}]{Sjoberg:1995}
Sj\"{o}berg, J., Zhang, Q., Ljung, L., A.~Benveniste, B.D., Glorennec, P.,
  Hjalmarsson, H., and Juditsky, A. (1995).
\newblock Nonlinear black-box modeling in system identification: A unified
  overview.
\newblock \emph{Automatica}, 31(12), 1691--1724.

\bibitem[{{Spinelli} et~al.(2005){Spinelli}, {Piroddi}, and
  {Lovera}}]{SPL_system}
{Spinelli}, W., {Piroddi}, L., and {Lovera}, M. (2005).
\newblock On the role of prefiltering in nonlinear system identification.
\newblock \emph{IEEE Transactions on Automatic Control}, 50(10), 1597--1602.
\newblock \doi{10.1109/TAC.2005.856655}.

\bibitem[{{Stoddard} et~al.(2017){Stoddard}, {Welsh}, and
  {Hjalmarsson}}]{Stoddard2017}
{Stoddard}, J.G., {Welsh}, J.S., and {Hjalmarsson}, H. (2017).
\newblock Em-based hyperparameter optimization for regularized volterra kernel
  estimation.
\newblock \emph{IEEE Control Systems Letters}, 1(2), 388--393.

\bibitem[{Tibshirani(1996)}]{Lasso1996}
Tibshirani, R. (1996).
\newblock Regression shrinkage and selection via the {LASSO}.
\newblock \emph{Journal of the Royal Statistical Society, Series B.}, 58,
  267--288.

\bibitem[{{Wigren} and {Schoukens}(2013)}]{silverbox}
{Wigren}, T. and {Schoukens}, J. (2013).
\newblock Three free data sets for development and benchmarking in nonlinear
  system identification.
\newblock In \emph{2013 European Control Conference (ECC)}, 2933--2938.
\newblock \doi{10.23919/ECC.2013.6669201}.

\end{thebibliography}

\end{document}